\documentclass{article}

\usepackage[nonatbib, final]{cml4impact_2022}

\usepackage[utf8]{inputenc} 
\usepackage[T1]{fontenc}    
\usepackage{hyperref}       
\usepackage{url}            
\usepackage{booktabs}       
\usepackage{amsfonts}       
\usepackage{nicefrac}       
\usepackage{microtype}      
\usepackage{xcolor}         
\usepackage{graphicx}
\usepackage{caption}
\usepackage{subcaption}
\usepackage{amsmath}

\title{Initial Results for Pairwise Causal Discovery Using Quantitative Information Flow}

\author{
    Felipe Giori \\
    Universidade Federal de Minas Gerais\\
    \texttt{felipegiori@dcc.ufmg.br} \\
  \And
    Flavio Figueiredo \\
    Universidade Federal de Minas Gerais\\
    \texttt{flaviovdf@dcc.ufmg.br} \\
}

\begin{document}

\maketitle

\begin{abstract}
Pairwise Causal Discovery is the task of determining causal, anticausal, confounded or independence relationships from pairs of variables. Over the last few years, this challenging task has promoted not only the discovery of novel machine learning models aimed at solving the task, but also discussions on how learning the causal direction of variables may benefit machine learning overall. In this paper, we show that Quantitative Information Flow (QIF), a measure usually employed for measuring leakages of information from a system to an attacker, shows promising results as features for the task. In particular, experiments with real-world datasets indicate that QIF is statistically tied to the state of the art.  Our initial results motivate further inquiries on how QIF relates to causality and what are its limitations.
\end{abstract}

\section{Introduction}

It is well known that the concept of causality has been studied for centuries, stimulating discussions among philosophers, legislators, economists, and computer scientists. One of the major objectives of many studies is to estimate cause-and-effect relationships based on observational data. For instance, using a sampled dataset are we able to determine that smoking causes lung cancer? Or that, studying at a high prestige university implies a higher salary?

In this paper, we focus on the task of finding causal relationships between pairs of variables. As it was shown by \cite{10.5555/3042573.3042635}, extracting such relationships is not only useful from a data exploration perspective, but it may also help us understand some of the limitations of machine learning models. That is, knowing the causal relationships between dependent and independent variables may help in choosing a good machine learning approach (e.g., semi-supervised vs supervised learning).

The classic example of this setting may be phrased from the perspective of the owner of an ice cream shop during the summer. In this season, such a person may notice both an increase in electricity bills as well as sales. Although there is a strong association between both variables, the weather acts as a confounding variable being the most likely cause of both other. Thus, when discovering cause and effect relationships between pairs of variables we must differentiate four scenarios. In other, are we able state that, (1) variable $X$ causes $Y$, $X \rightarrow Y$, or that (2) $Y$ causes $X$, $Y \rightarrow X$. If neither is the case then either, (3) $X$ and $Y$ are related through some confounding effect $Z$, $X \leftarrow Z \rightarrow Y$, or (4) the variables are independent, i.e., $X \perp Y$.


Motivated by the importance of the problem and its challenges, this paper focuses on exploring novel features for causal discovery extracted using Quantitative Information Flow (QIF) \cite{Alvim_Chatzikokolakis_McIver_Morgan_Palamidessi_Smith_2020}. QIF is an information-theoretic framework that allows us to measure how much information flows through a system. Originally, QIF was developed to measure information leakage in systems for security purposes. This is done by using observable outputs of a channel or function to update our beliefs about the channel input. To understand QIF, imagine you want to guess a 4-PIN password. There are 10,000 possible combinations for that password. When we enter a randomly chosen password in the password-checker, we will likely get a message saying that the password is incorrect. In this case, the password is the input, the password-checker is the channel and the message is the output. By observing the channel output, we gain some information about the input, that is, we know that the password we typed was incorrect and now we have "only" 9,999 possibilities left. This means that the channel leaked some information about the true password. QIF is a framework employed to measure how much information is leaked based o different attack strategies.

When approaching the causal discovery task, evaluations are performed in manually labeled real-world datasets \cite{10.5555/2946645.2946677} or synthetic pairs of variables\footnote{http://www.causality.inf.ethz.ch/cause-effect.php}. Whereas some datasets include all of the four scenarios above, others are restricted to defining causal ($X \rightarrow Y$) and anticausal ($Y \rightarrow X$) relationships. Currently, one of the most promising approaches to differentiate across the four relationships above is Jarfo \cite{Fonollosa2019} which computes several statistical measures of the cause-effect pairs and trains a GradientBoosting model to predict the causal direction. As we show in this paper, not only are QIF features, by themselves, able to proving promising results for caual discovery, combining the features computed by the QIF framework with the Jarfo features yields the best results in most of our tests. Overall, we also show that using the QIF features are a promising direction that needs further exploration from the ML community.

The rest of this paper is organized as follows, in Section 2 we go over the related work and give a brief overview of the QIF framework, in Section 3 we present our experiments and results and Section 4 we discuss our results and propose future work.

\section{Background and Related Work}

Causality is the influence some set of events, or random variables, named causes have on the outcome of other variables, the effects. There are several ways to model cause-effect relationships, but in this paper, we will use causal graphical models (CGMs) pioneered by \cite{Spirtes2000} and \cite{10.5555/1642718}, and functional causal models (also known as structural causal models). 
In the domain of machine learning, we can categorize previous work on CGMs into two overlapping sub-domains: causal inference and causal discovery. In causal inference, the aim is to predict how changes on the cause variables affect the response variables. Currently, this is mostly done via the do-calculus \cite{10.5555/1642718}. 
In contrast, causal discovery is focused on identifying cause and effect relationships from observational data. 

\subsection{Causal Discovery}

It is well known that general causal discovery, that is, uncovering full CGMs from several random variables from observation alone is in general not possible. However, in some limited settings and under some assumptions this is possible and previous endeavors have shown promising results. Such efforts may be categorized as: constraint-based, score-based and methods based on algorithms that exploit functional causal models.

\textbf{Constraint-based algorithms} focus on learning a set of causal graphs that satisfy the conditional independence in the data. A famous algorithm is the Peter-Clark Algorithm \cite{doi:10.1177/089443939100900106} and its variant introduced by \cite{10.5555/1248659.1248681}. This algorithm starts with a fully connected graph and tests for conditional independence in order to select which edges to remove. The fast causal inference algorithm \cite{Spirtes2000} and its extensions presented by \cite{10.1214/11-AOS940} and \cite{10.5555/2074158.2074215} take unobserved confounders into account when generating the graph.

\textbf{Score-based algorithms} replace the conditional independence tests with the goodness of fit tests, that is, we want to maximize some score criterion. The Bayesian Information Criterion (BIC) score \cite{10.1214/aos/1176344136} is a widely adopted metric, defined as:
    $S(X, G') = log P(X|\hat{\theta}, G') - \frac{J}{2}log(n)$.
\noindent Where, $G'$ is a causal graph, $X$ is the data, $\hat{\theta}$ is the maximum likelihood estimator of the parameters, $J$ is the number of variables and $n$ the number of instances. Searching for a causal-graph with maximum score is a NP-Complete problem, as shown by \cite{DBLP:conf/aistats/Chickering95}. 

\textbf{Algorithms based on functional causal model} exploit the function $y = f(x, \epsilon)$, where $y$ is the effect, $x$ is the cause and $\epsilon$ is some noise term. A famous algorithm is the LiNGAM introduced by \cite{10.5555/1248547.1248619} where we try to estimate some lower triangle matrix $A$ such that $x = xA + \epsilon$. There are variations of the LiNGAM algorithm such as ICA-LiNGAM~\cite{10.5555/1248547.1248619} and DirectICA-LinGAM~\cite{10.5555/1953048.2021040}



\subsection{Causal and Anticausal Machine Learning}

One of the goals of machine learning is to exploit statistical associations or dependence between endogenous variables to predict the value of other, exogenous, variables. Endogenous variables are usually defined as $X_i$, with $i$ being some arbitrary index, whereas the exogenous responses are $Y_j$. For simplicity, let us focus on the pairwise setting, where we only have one $X$ and one $Y$. In this setting, the goal of supervised machine learning algorithms is to estimate some model, or hypothesis, $h_\theta: X \rightarrow Y$. Here, $\theta$ captures a vector of model parameters, with one example of such a model being a simple logistic regression model, i.e., $y = h_\theta(x) = \frac{1}{1 + e^{-(\theta_0 + \theta_1 x})}$. 

Within this supervised learning, researchers have defined what is known as causal and anticausal learning. Consider the setting where we are trying to predict whether one has lung cancer, $Y$, based on the average amount of cigarettes one smokes per year. It is well known that smoking causes cancer, thus some real causal relationship $f:X \rightarrow Y$ exists and we are trying to estimate causes from effects $h_\theta: X \rightarrow Y$. Now, consider a similar setting where we are still trying to predict lung cancer, but from some x-ray image. Here, $X$ is a feature associated with the number of dark pixels in this image (usually related to cancer). There is no way some computer image caused cancer on an individual, here, the image is an effect of having cancer, i.e., $f:Y \rightarrow X$ (the opposite of the previous relationship). Nevertheless, via machine learning, we still aim to estimate $h_\theta: X \rightarrow Y$, thus capturing an anticausal learning task.



Several papers have explored the causal or anticausal nature of supervised learning in different settings. For instance, \cite{10.5555/3042573.3042635} showed semi-supervised learning is usually only beneficial in the causal direction. Moreover, \cite{DBLP:journals/corr/abs-1802-03916} explored anticausal learning in order to estimate the degeneracy of machine learning models over time.




\subsection{Quantitative Information Flow} \label{qif}

Quantitative Information Flow (QIF) is an information theoretic framework that computes how much information flows through a system. QIF was developed to study information security from a perspective of attacks to a system. In order to understand QIF, consider the setting where an adversary wants to know the value of some secret $x$ in the domain of the random variable $X$.
Given that the attacker want's to guess $x$, she/he may use some prior knowledge to guess this value. This prior knowledge is determined by some prior probability mass function: $\pi_x$. One possible, smart, strategy 
would be to choose the most likely value of $x$, that is, the value with maximum probability. The intuition is that if we only have one shot to guess the value $x$, then it is smart to choose the most likely value. We call this the Bayes Vulnerability, defined as:
\begin{equation}\label{bayes_vulnerability}
    V_b(\pi) = \max_{x \in X}\pi_x.
\end{equation}
Bayes vulnerability measures one example of a strategy an attacker may take, there are other measures that capture other definitions of information flow. For instance, Shannon's Entropy can be viewed as a measure that captures an attacker guessing random secrets sampled according to $\pi$. Regardless of the choice of attack, the work by \cite{10.1109/CSF.2012.26} showed that we measure the flow of information flow from system to attacker. The intuition is that for every possible action an adversary can perform, there is a corresponding gain. That is, for every action $w \in W$ that the adversary can perform, there is some associated gain for every $x \in X$. This gain is defined by some gain function $g:W\times X \rightarrow \mathbb{R}$ where an action $w \in W$ can be seen as a guess that the adversary could make about the secret, in which case it is natural that $W = X$. For example, in the password checker problem, an action could be to guess password $a$ and the gain function then maps a gain for every value of $x \in X$ if we had taken that action. We can also think of an action as not making a guess if we think the situation is too risky. Now we can measure the threat of some adversary towards a secret as the expected gain for a optimal choice of $w \in W$. This is a prior g-vulnerability, defined as
\begin{equation}
    V_g(\pi) = \sup_{w\in W} \sum_{x \in X} \pi_x g(w, x).
\end{equation}
Now, lets consider a channel $C$ that processes a secret $x \in X$. A channel $C$ is a function $C:X \rightarrow \mathbb{D}Y$, so that for each $x$ in $X$ , we have $C(x)$ as a distribution on $Y$, giving the probability of each possible output value $y$. Entry $C(x,y)$ denotes $P(y|x)$, the conditional probability of getting output $y$ given input $x$. When a secret passes through a channel and output $y$ is observed, some information might be released, that is, observing output $y$ of a channel might help us increase our knowledge about secret $x$. We can now compute the threat represent by the adversary given that the output was observed. We call this new threat the posterior g-vulnerability, defined as
\begin{equation}
    V_g(\pi \triangleright C) = \sum_{y \in Y} \sup_{w \in W} \sum_{x \in X} \pi_x C(x,y)g(w,x).
\end{equation}
Given both prior and posterior g-vulnerabilities, we can measure how much knowing $y$ increases our knowledge about secret $x$ by comparing the computed vulnerabilities. This measure is called g-leakage and can be computed in a additive or multiplicative manner, as follows:
\begin{equation}
    \mathcal{L}^\times_g (\pi \triangleright C) = \frac{V_g(\pi \triangleright C)}{V_g(\pi)}
\end{equation}
\begin{equation}
    \mathcal{L}^+_g (\pi \triangleright C) = V_g(\pi \triangleright C) - V_g(\pi)
\end{equation}
The leakage functions return how much more vulnerable the secret is once the output $y$ is observed. As shown by \cite{thiago2019} we can compute the information flow in both directions, from $X$ to $Y$, called direct flow, and from $Y$ to $X$, called reverse flow. In our experiments we used the QIF framework to compute leakages in both directions and used those as variables to train classifiers to predict the causal direction. We used the identity gain function defined as:
\begin{equation}
g_{id}(x, w) = 
\left\{
    \begin{array}{lr}
        1, \text{if } w = x\\
        0, \text{if } w \neq x
    \end{array}
\right\}
\end{equation}
to compute the Bayes Vulnerability and used an alternate version of the framework that, instead of maximizing a gain function, minimizes a loss function. We used the following loss function to measure the Shannon leakage:
\begin{equation}
    \ell_{H}(w, x) = -\log_2 w_x.
\end{equation}
Finally, we also measured the channel capacity which is a measure of how much information the channel can leak independent of the prior and the gain function.

\section{Experiments and Results}

In this section, we provide our experimental setup and share some of the results obtained with models trained using variables computed by the QIF framework.

\textbf{Dataset}: In our experiments, we use two datasets. The first is the cause-effect pairs from the cause-effect pair challenge\footnote{http://www.causality.inf.ethz.ch/cause-effect.php} consisting of 20k datasets of synthetic pairs of variables labeled as causal, anticausal, confounded, and independent. The second dataset consists of 99 real-world known cause-effect pairs made available by \cite{10.5555/2946645.2946677}. This second set of data is called the Tuebingen dataset. The dataset simply states the causal direction of variables, thus we randomized the pairs in order to have two labels, causal and anticausal, of similar sizes (roughly 50 pairs).

\textbf{Experimental setup}: Our approach is similar to \cite{Fonollosa2019}  where we compute several statistical measures from the relationship between the cause-effect pairs and train a machine learning model on those statistics to predict the causal direction. To serve as a baseline we train a model on the features computed by Jarfo \cite{Fonollosa2019} and compare the results of a model trained with features computed by QIF. We also consider a third setting where we combine features from both approaches. We also point out that we trained classifiers for solving two problems: predicting causal or anticausal and predicting all four classes on the cause-effect pair challenge datasets. Finally, given that labels are of the same size, we measure accuracy as a metric of performance (results are reported with 95\% confidence intervals). 

\textbf{Model training}: We tested different learning methods for classification and, on average, we found the best results using the XGBoost classifier. We perform a 10-fold cross-validation in all of our experiments and we test our predictions on a holdout test set containing 20\% of the data.

\textbf{QIF features}: The QIF framework implementation we used has limitations where it needs relative frequencies to measure the leakages. These relative frequencies can be quite tricky to compute when we have continuous variables. In order to overcome these limitations, we adapted the framework to use a Gaussian kernel estimated by the numerical variables pairs and use this kernel to compute the leakages. When we had a numerical variable and another non-numerical variable we used a simple binning discretizer and optimized the number of bins in cross-validation. Finally, the features we used were the Bayes vulnerability, risk, and Shannon leakages, such as multiplicative and additive, the difference between direct and reverse leakages of those measures, and the channel capacity.

\begin{figure}[ht]
     \centering
     \begin{subfigure}[t]{0.24\textwidth}
         \centering
         \includegraphics[width=\textwidth]{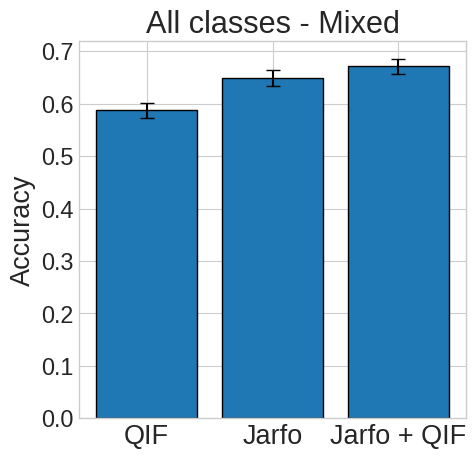}
         \caption{All classes and all variable types}
         \label{fig:all_classes_mixed}
     \end{subfigure}
     \hfill
     \begin{subfigure}[t]{0.24\textwidth}
         \centering
         \includegraphics[width=\textwidth]{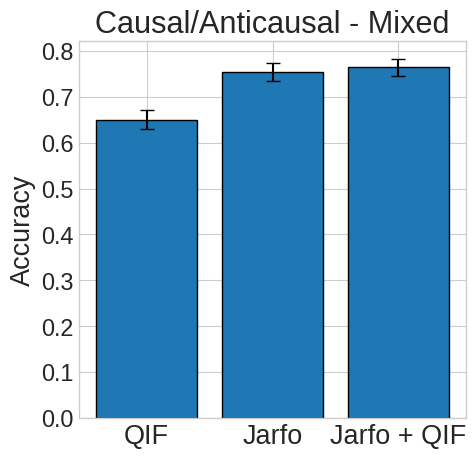}
         \caption{Causal/Anticausal and all variable types}
         \label{fig:causalanticausal_mixed}
     \end{subfigure}
     \hfill
     \begin{subfigure}[t]{0.24\textwidth}
         \centering
         \includegraphics[width=\textwidth]{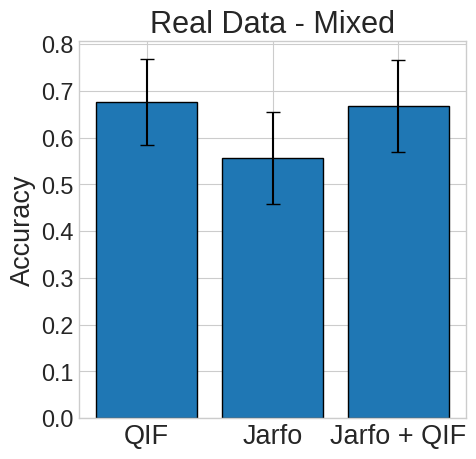}
         \caption{Model trained on all variable types of the challenge cause-effect pairs and tested on real data}
         \label{fig:real_data_mixed}
     \end{subfigure}
     \hfill
     \begin{subfigure}[t]{0.24\textwidth}
         \centering
         \includegraphics[width=\textwidth]{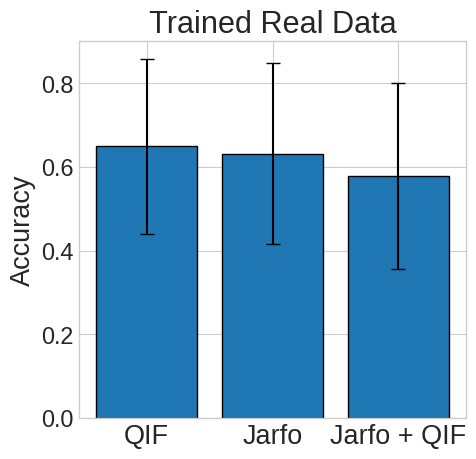}
         \caption{Accuracy results with the model trained and tested on real data}
         \label{fig:real_data}
     \end{subfigure}
        \caption{Accuracy results for the challenge cause-effect pairs}
        \label{fig:accuracy_mixed}
\end{figure}

In our first experiment, we trained models on the challenge dataset. Given that in this dataset we have all four classes, we present results for the four classes as well as for the causal vs anticausal setting only. The results are shown in Figure \ref{fig:accuracy_mixed}. From the figure, we can see that in the problem where we have to predict all four classes, the classifiers' performances were quite similar, with the best results found in the classifier using both QIF and Jarfo features. In the causal/anticausal setting, the model using only QIF features performed significantly below the other two. However, it is important to point out that in real-world datasets, using only QIF is as good as using several other features explored by Jarfo. These numbers are shown in Figure \ref{fig:real_data}.

\begin{figure}[ht]
     \centering
     \begin{subfigure}[t]{0.49\textwidth}
         \centering
         \includegraphics[width=\textwidth]{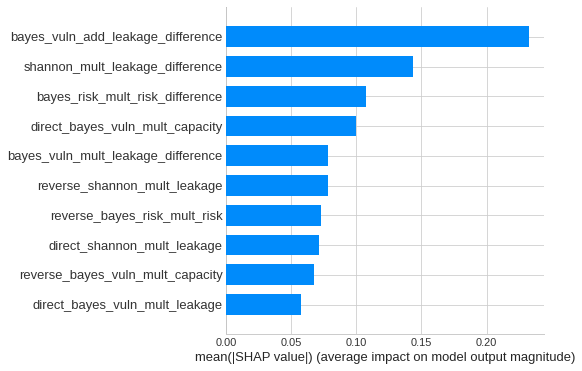}
         \caption{Feature importance of the top 10 features of the model trained only with QIF variables for the causal/anticausal problem}
         \label{fig:feature_importance}
     \end{subfigure}
     \hfill
     \begin{subfigure}[t]{0.49\textwidth}
         \centering
         \includegraphics[width=\textwidth]{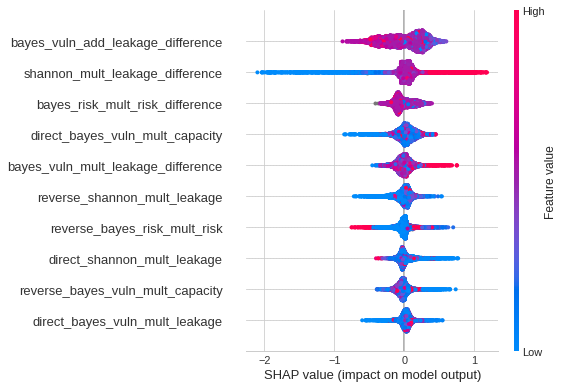}
         \caption{SHAP values of the top 10 features of the model trained only with QIF variables for the causal/anticausal problem}
         \label{fig:shap_values}
     \end{subfigure}
        \caption{Feature importance and SHAP values}
        \label{fig:importance_shap}
\end{figure}

We computed the feature importance and Shapley additive explanations \cite{NIPS2017_7062}, which can be seen in figure \ref{fig:importance_shap}. The most important features were the difference between direct and reverse leakages for the Bayes vulnerability, Shannon, and Bayes risk. A clear relationship on the Shannon leakage, where high values have a positive impact and low values have a negative impact. The Bayes vulnerability additive leakage difference also shows a moderate relationship where low values have a positive impact and high values have a negative impact. That means that when the reverse leakage is higher, we tend to predict causal orientation and when the direct leakage is higher, we tend to predict anticausal orientation.

Our results in these initial experiments show that QIF measures are a prospective line of research for the causal discovery task. Our goal in future experiments is to better understand how QIF relates, in a theoretical sense, to other causal discovery approaches discussed in Section 2.

\section{Conclusion and Future Work}
In this paper, we studied the impact of trying to predict the causal direction of the cause-effect pairs by using models trained on features computed by the quantitative information flow framework. We used the features computed by Jarfo as a baseline and saw that our results were slightly better on the real-world cause-effect pairs. 

Initial promising lines of future work may be performed by expanding the set of gain functions used by QIF. This allows one to compute other features which may improve results. 

Before concluding, it is important to point out that we did experiment with different approaches to estimate probability distributions for numeric datasets before using QIF (e.g., Kernel Density Estimators \cite{Scott1992} or Flow Models \cite{nflows} \cite{NEURIPS2019_7ac71d43}). In these settings, we found that results are mostly the same as the binned categorization we performed. Thus, we believe that the QIF, as is, is a promising approach for causal discovery and needs to be better understood theoretically (e.g., how the measure relates to other causal discovery approaches such as LiNGAM or ANM).


\section*{Acknowledgements}

This work has been partially supported by the project GO Deep (Geo-sciences Oriented Deep Learning) project funded by Petrobras (2019/00480-9  	PT-200.20.00121). Funding was provided by the authors’ individual grants from CNPq and  CAPES.  In particular, CNPQ's Universal 2018 Grant: 421884/2018-5.

\bibliographystyle{abbrv}
\bibliography{neurips_2022.bbl}

\end{document}